\documentclass[sigconf]{acmart}

\setcopyright{none}
\renewcommand\footnotetextcopyrightpermission[1]{}

\AtBeginDocument{%
  }

\usepackage{multirow}

\begin{document}

\title{Agent4cs: A Multi-agent System for Code Summarization in Large Hierarchical Codebases}


\author{Yongjian Tang}
\orcid{0009-0008-8670-6843}
\affiliation{%
  \institution{Siemens AG}
  \institution{Technical University of Munich}
  \city{Munich}
  \state{Bayern}
  \country{Germany}
}

\author{Ezgi Sarikayak}
\affiliation{%
  \institution{Siemens AG}
  \institution{Technical University of Munich}
  \city{Munich}
  \state{Bayern}
  \country{Germany}
}

\author{Doruk Tuncel}
\affiliation{%
  \institution{Siemens AG}
  \city{Munich}
  \state{Bayern}
  \country{Germany}
}

\author{Jie M. Zhang}
\affiliation{%
 \institution{Kings College London}
 \city{London}
 \country{the United Kingdom}
}

\author{Thomas Runkler}
\affiliation{%
  \institution{Siemens AG}
  \institution{Technical University of Munich}
  \city{Munich}
  \state{Bayern}
  \country{Germany}
}

\renewcommand{\shortauthors}{Tang et al.}

\begin{abstract}
Understanding large, complex codebases, especially those with obfuscated structures and incomplete documentation, remains a significant challenge. Existing code summarization solutions often rely on a single language model or coding assistant like Claude Code, and treat source code as flat text, underutilizing the rich interdependencies and hierarchical information within a repository. To address these shortcomings, we propose Agent4cs -- a multi-agent framework that summarizes large codebases in a bottom-up fashion, where a summarization agent focuses on producing robust summaries; a keyword-extraction agent proactively identifies critical information from subfolders; and a quality-assurance agent iteratively refines the outputs for readability, coherence, and completeness. Evaluated on 7 frontier models, Agent4cs improves semantic consistency across all folder levels by average 8\% compared to two structured prompting baselines with code segments. Furthermore, extensive evaluation on real-world datasets demonstrates up to 38\% gains in normalized keyword coverage rate over the same baselines.
\end{abstract}




\maketitle

\pagestyle{plain}

\section{Introduction}
As software projects evolve rapidly, they often lead to incomplete, outdated, or inconsistent documentation. This friction slows on-boarding, complicates maintenance, and increases the risk of defects and architectural drift. To address these problems, considerable efforts have been dedicated to automated code summarization, such as identifying class stereotypes and retrieving functional keywords \cite{zhu2019automatic}. More recently, deep learning approaches have advanced this field by utilizing neural models to learn complex mappings between code structures and natural language descriptions \cite{zhang2022survey,shi2022evaluation}. The advent of Large Language Models (LLMs) \cite{sun2025source,zhang2024review} has further minimized the semantic gap between code and natural language, offering new avenues for robust code summarization.

While advances in code-aware language models have significantly improved function-level summarization \cite{sun2025source}, achieving comprehensive repository-level summaries remains challenging -- cross-file dependencies and hierarchical information are often dispersed across many directories and therefore not fully utilized.  Prior efforts have shown that simply prompting an LLM is inadequate to capture all sub-modules throughout the repository \cite{dhulshette2025hierarchical,makharev2025code,hanya}, and interactive agents like Claude Code are designed for on-demand query-based code exploration rather than generating persistent documentation in a systematic way \cite{tang2026concept}. This problem is amplified for industrial codebases that often far exceed 300K tokens. Therefore, a more structured solution is required.

\begin{figure*}[h]
  \centering
  \includegraphics[width=1\linewidth]{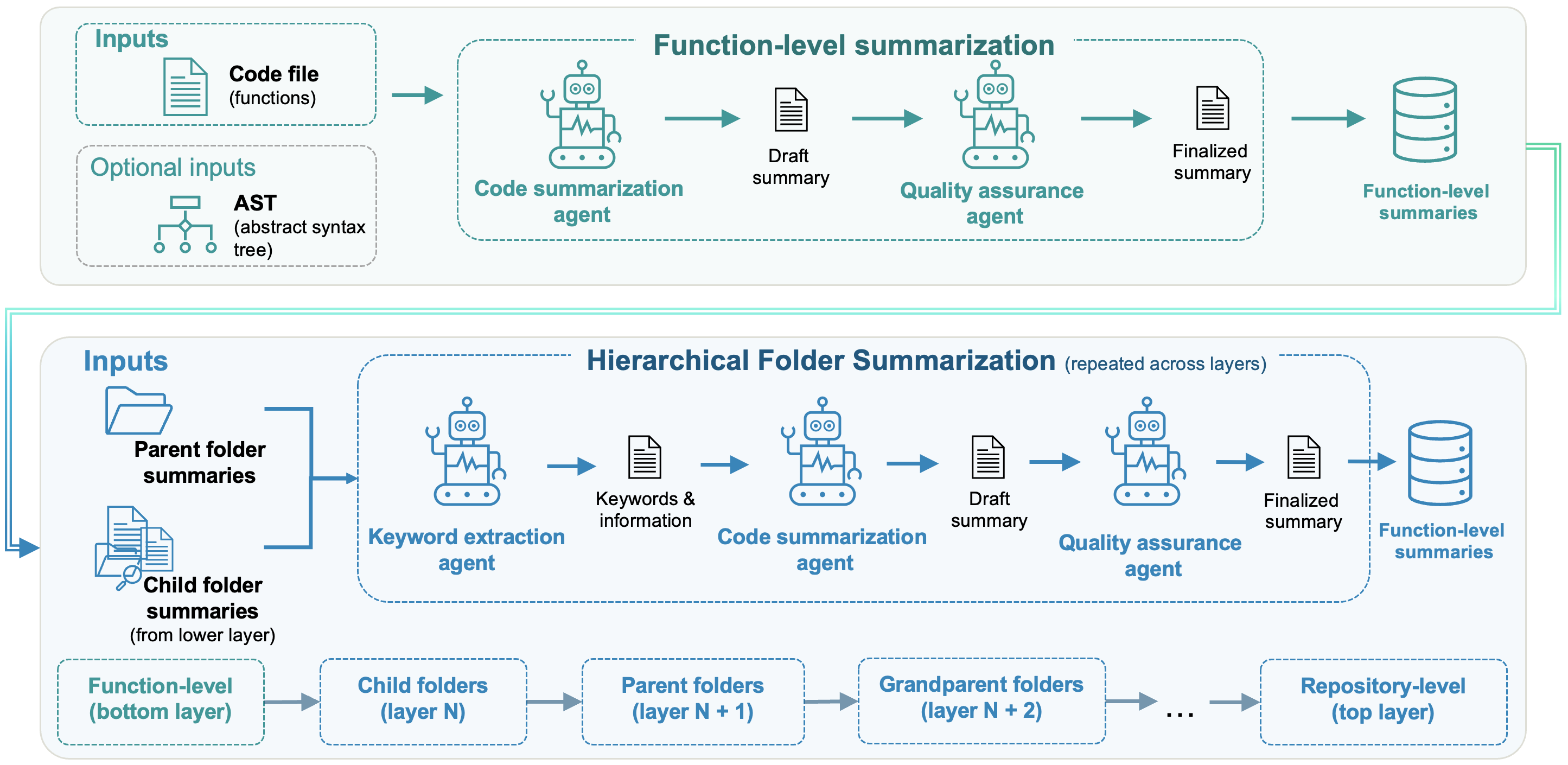}
  \caption{
  Agent4cs performs hierarchical repository summarization in two stages: function-level summarization and hierarchical folder-level summarization.}
  \label{fig:agent_overview}
\end{figure*}

To mitigate this gap, we propose Agent4cs -- an agentic framework that features a keyword extraction agent to capture latent information from subfolders and a quality assurance agent to provide iterative feedback for summary refinement, as illustrated in Figure \ref{fig:agent_overview}. Combined with a bottom-up approach for hierarchical code summarization, the framework strengthens the cross-folder connections within repositories while ensuring summary quality and readability, serving as a ideal complementary tool for existing coding agents. 
The main contributions of this work can be summarized as follows: 

\begin{itemize}
    \item [(i)]  
    We propose Agent4cs -- a lightweight agentic framework for repository-level code summarization. To the best of our knowledge, this represents the first multi-agent approach for hierarchical codebase summarization.

    \item [(ii)] 
    Following a bottom-up summarization strategy, we effectively leverage hierarchical information within repositories to enhance cross-folder connections, achieving an 8\% average improvement in semantic similarity between folder and subfolder summaries compared to two structured prompting baselines that incorporate segment-level code context.

    \item [(iii)] 
    We analyze the performance of Agent4cs by integrating seven frontier LLMs spanning different performance tiers, studying its effectiveness and robustness across diverse model capabilities and computational budgets.

    \item [(iv)] 
    We validate our approach on filtered data from three real-world datasets and their obfuscated versions, surpassing the code segment-augmented baselines in normalized keyword coverage rate by approximately 38\% with certain LLMs while maintaining summary length. Meanwhile, we improve the readability of generated summaries across most evaluated LLMs.
\end{itemize} 

\section{Problem Statement}
\begin{figure}[h]
  \centering
  \includegraphics[width=1\linewidth]{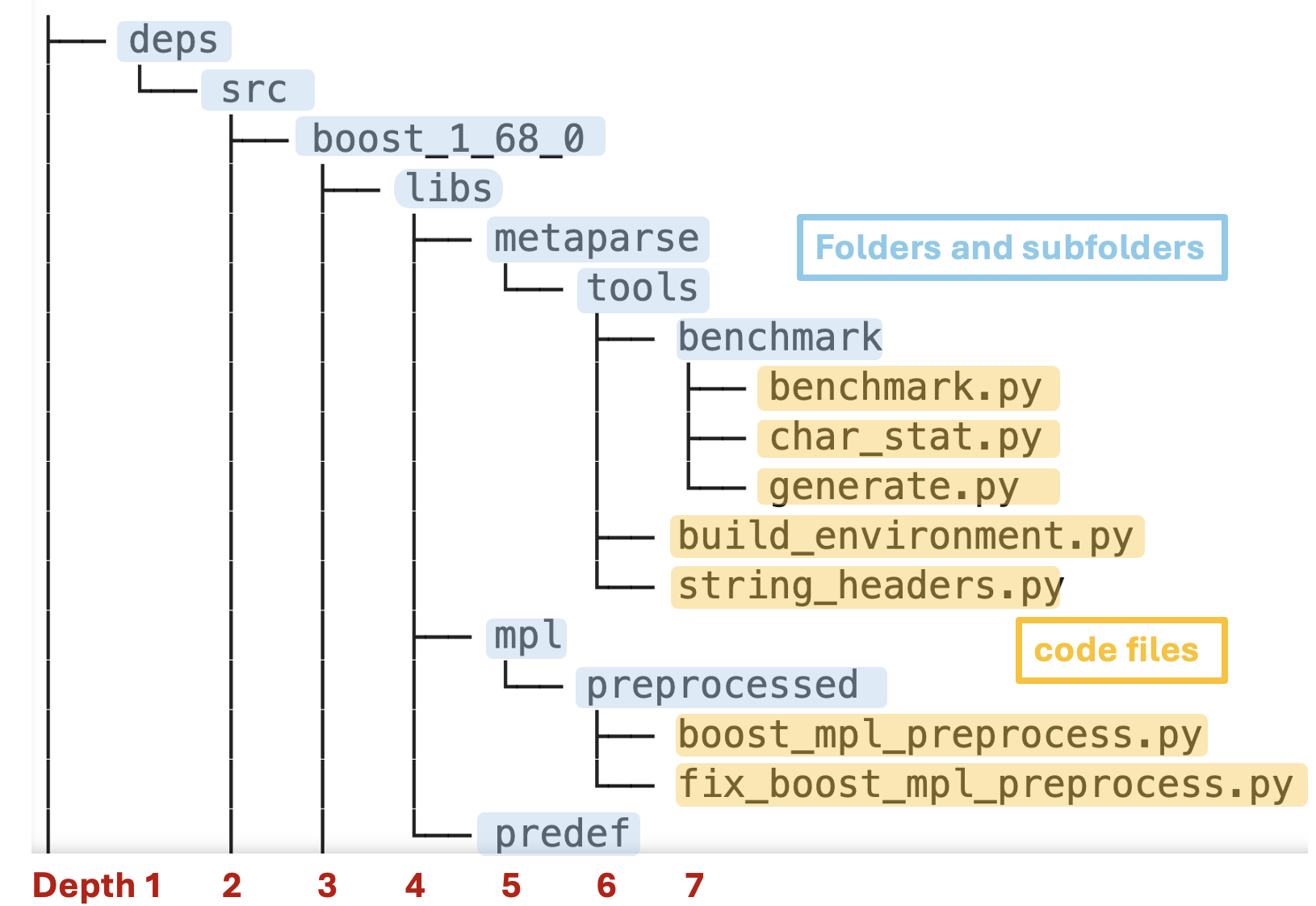}
  \caption{Overview of a repository: hierarchical folder structures (in blue) and function-level code files (in yellow).}
  \label{fig:problem_statement}
  \Description{Overview of a repository: hierarchical folder structures (in blue) and function-level code files (in yellow).}
\end{figure}

Figure \ref{fig:problem_statement} exemplifies a representative subset of a large Python repository, showcasing its multi-folder architecture and source code files in the bottom layer. For clarity, we divide code summarization into two tasks: function-level summarization for individual code files and hierarchical summarization for repository-wide understanding \cite{zhang2022survey, zhu2019automatic}. 

\subsection{Function-level Code Summarization}
The code files (yellow) in Figure \ref{fig:problem_statement} contain code snippets and functions. The groundtruth data from most established datasets applies to this problem scope, providing human-written function-summary pairs as evaluation benchmarks. \sloppy{Nevertheless}, this setting is constrained to local code interpretation. For repository-level understanding, such function-oriented benchmarks fall short.

\begin{figure}[htbp]
  \centering
  \includegraphics[width=0.9\linewidth]{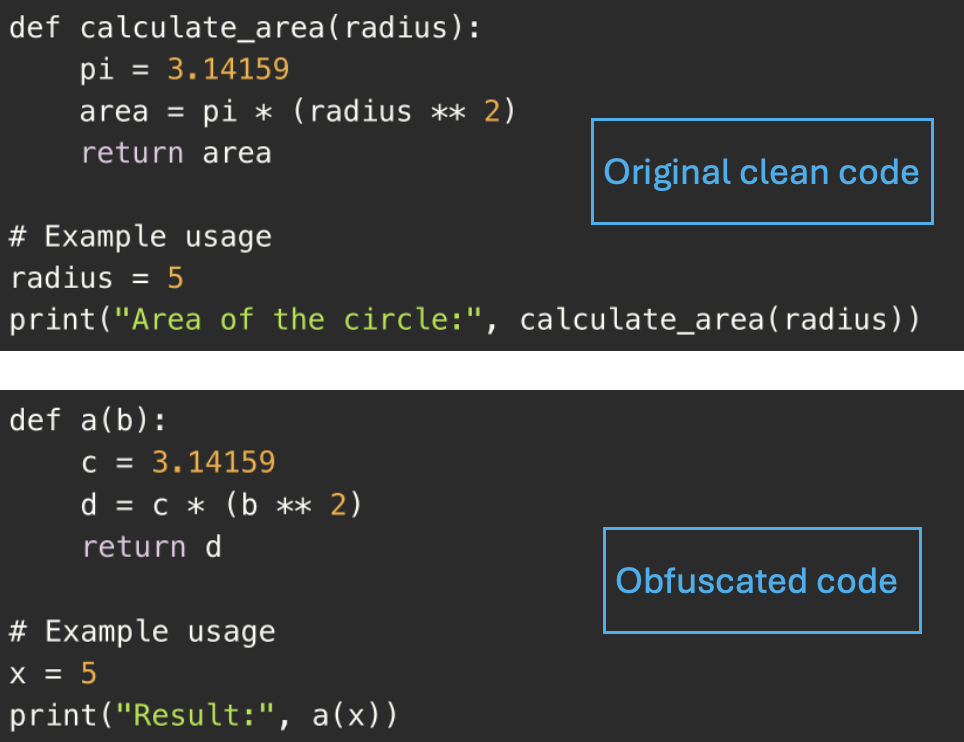}
  \caption{An example of code obfuscation through identifier renaming.}
  \label{fig:obfuscated_code_example}
  \Description{An obfuscation example of identifier renaming.}
\end{figure}

In our experiments, we extend beyond standard clean-code datasets and include obfuscated code to increase the difficulty and simulate real-world scenarios. Code obfuscation \cite{sebastian2016study,viticchie2016assessment} is a deliberate transformation strategy that conceals the logic and structure of source code while preserving its functionality, widely employed in industry to protect intellectual property and prevent malicious code analysis. Common obfuscation techniques include structural modification, identifier renaming, structural modification, and control flow alteration, all designed to make code difficult to understand and analyze. Figure \ref{fig:obfuscated_code_example} demonstrates an example of obfuscated code, where original identifiers (\texttt{calculate\_area}, \texttt{radius}, \texttt{pi}, \texttt{area}) were replaced with arbitrary, non-descriptive names (\texttt{a}, \texttt{b}, \texttt{c}, \texttt{d}, \texttt{e}), which reduces semantic clarity and makes the code more challenging to interpret. 

Before LLMs, traditional summarization approaches often struggle with obfuscated code because the semantic meaning is hidden beneath layers of syntactic transformation. 
Evaluating LLMs on obfuscated code provides insights into model robustness and their ability to extract underlying functionality despite obfuscation, making this valuable for real-world code understanding scenarios.

\subsection{Hierarchical Code Summarization}
Most existing studies overlook summaries of higher-level folders and subfolders because available datasets, collected from open-source repositories, lack ground-truth annotations at this level. \sloppy{However}, as modern software projects continue to grow in complexity and scale, understanding code at folder and repository levels becomes increasingly critical for effective long-term maintenance. Consequently, we identify hierarchical code summarization as a distinct problem within the broader code summarization domain and propose novel solutions and evaluation metrics to address this challenge in this work.

\section{Related Work}


\subsection{Neural Models for Code Summarization}

Motivated by the success of neural machine translation in natural language processing \cite{bahdanau2014neural, cho-etal-2014-properties}, encoder-decoder architectures were soon adapted for code summarization. Early work like CODE-NN trained an end-to-end LSTM 
to generate summaries for C\# and SQL queries \cite{iyer2016summarizing}. Building upon these sequential neural models, subsequent research utilized structural code information by additionally incorporating Abstract Syntax Tree (AST) alongside code snippets into 
deep reinforcement learning frameworks \cite{wan2018improving}. This structural awareness evolved further with graph-based neural architectures, which moved beyond flattened AST representations to preserve the hierarchical nature of AST trees \cite{leclair2020improved}.  
Several follow-up studies combined neural models with retrieval techniques \cite{liu2021retrieval, zhang2020retrieval}, enhancing encoder-decoder architectures by leveraging syntactically and semantically similar code samples from training corpora.

\subsection{Language Models for Code Summarization}
The emergence of transformer-based language models \cite{vaswani2017attention} marked a paradigm shift in code summarization, moving from task-specific neural architectures to fine-tuning pre-trained models on code-related tasks. These attention-based models have demonstrated superior performance in capturing long-range dependencies and 
significantly outperformed neural approaches in summarizing source code \cite{ahmad2020transformer}. To investigate whether these models truly understand code semantics or merely rely on surface-level textual cues, researchers modified function and variable names and found that such changes had minimal impact on model performance \cite{mondal2023understanding}. Meanwhile, concerns about resource consumption have driven the development of more compact alternatives, such as distilling GPT-3.5's code expertise into a model with 350M parameters\cite{su2024distilled}. Building on these foundations, several innovative approaches have emerged. Inspired by genetic algorithms, the EACS framework improves candidate summaries using selection, crossover, and mutation operations, which consistently improves the quality of code documentation across large software projects \cite{sun2024extractive}. Moreover, 
ESALE combined neural architecture with transformers and employed a two-phase approach -- pre-training a shared encoder via multi-task learning and then fine-tuning it with a decoder for task-specific code summarization \cite{fang2024esale}.

In line with the development of language modeling, solutions to code summarization have soon evolved from fine-tuning to efficient prompt-based inference with LLMs. Although early investigation of ChatGPT in code summarization observed inferior performance to fine-tuned language models \cite{sun2023automatic}, recent studies with more advanced SOTA models have revealed that LLMs using prompting strategies can outperform fine-tuned models in most code summarization and translation benchmarks \cite{shin2025prompt, sun2025source}. To understand what drives current LLMs' effectiveness in code summarization, researchers also examined the impact of token overlap between code and the corresponding natural language references, revealing that LLMs often rely more heavily on these shared tokens than on underlying code structure \cite{haldar2024analyzing}.
To strengthen LLMs' competence in code understanding and other domain-specific tasks, few-shot prompting with demonstration examples is widely used \cite{10.1109/ASE56229.2023.00109, 10.1145/3551349.3559555, ahmed2024automatic, tang2024fsponer}, which can outperform a fine-tuned model trained by thousands of samples using only several demonstration examples. Subsequent research has explored the upper limits of this technique by incorporating hundreds of semantically selected few-shot examples in software engineering domain \cite{tang2025few}.

\subsection{Hierarchical Code Summarization}
\label{subsection:selected baseline}
While LLM-based techniques have proven their efficiency for individual code snippets, the challenge of summarizing large codebases with complex interdependencies has driven researchers to develop approaches that can capture multi-layer code structures. 
A syntax-driven approach for Hierarchical Repository-Level Code Summarization (HR-CS) \cite{dhulshette2025hierarchical} employed local LLMs to aggregate code segment-level information such as function names, variables, inputs, and outputs, and then produce comprehensive package-level summaries for improved relevance and coverage. 
Another work extended Code Summarization Beyond Functions (CS-BF) by incorporating class and repository domain context and few-shot examples \cite{makharev2025code}. While this repository-level summarization shows promise, it remains compute-intensive and omits mid-level folder layers within repositories. By contrast, other studies utilize summarization as an intermediate step for downstream tasks. For instance, HCGS generated summaries of code elements and their relationships to achieve context-aware code retrieval \cite{sounthiraraj2025code}, while another method leveraged LLMs to build hierarchical project summaries for bug localization, performing top-down inference to overcome domain mismatch and context limits \cite{10.1007/978-3-031-97576-9_6}. 

In this work, we focus on hierarchical code summarization and select HR-CS and CS-BF as baselines to directly measure improvements in experiments.



\subsection{LLM-Based Evaluation}
Beyond generating summaries, recent research has explored leveraging language models as automated evaluators to judge the quality of code understanding and summarization.
CodeJudge-Eval establishes a benchmark for measuring LLMs' code understanding abilities through their performance in judging code solution correctness \cite{zhao2025codejudge}, while CODERPE integrates multi-role prompts into LLMs to automatically evaluate code summaries across coherence, fluency, and relevance \cite{wu2025can}. Despite these advances, empirical evidence shows that while large-scale models, such as GPT-4-turbo, exhibit promising capabilities in judging summary quality, smaller LLMs struggle, and even the best one frequently misjudges, indicating that robust LLM-based evaluation remains an open challenge \cite{crupi2025effectiveness}.
\section{Agent4cs}
\subsection{Agentic Workflow for Function-level Summarization}

\begin{figure}[h]
  \includegraphics[width=1\linewidth]{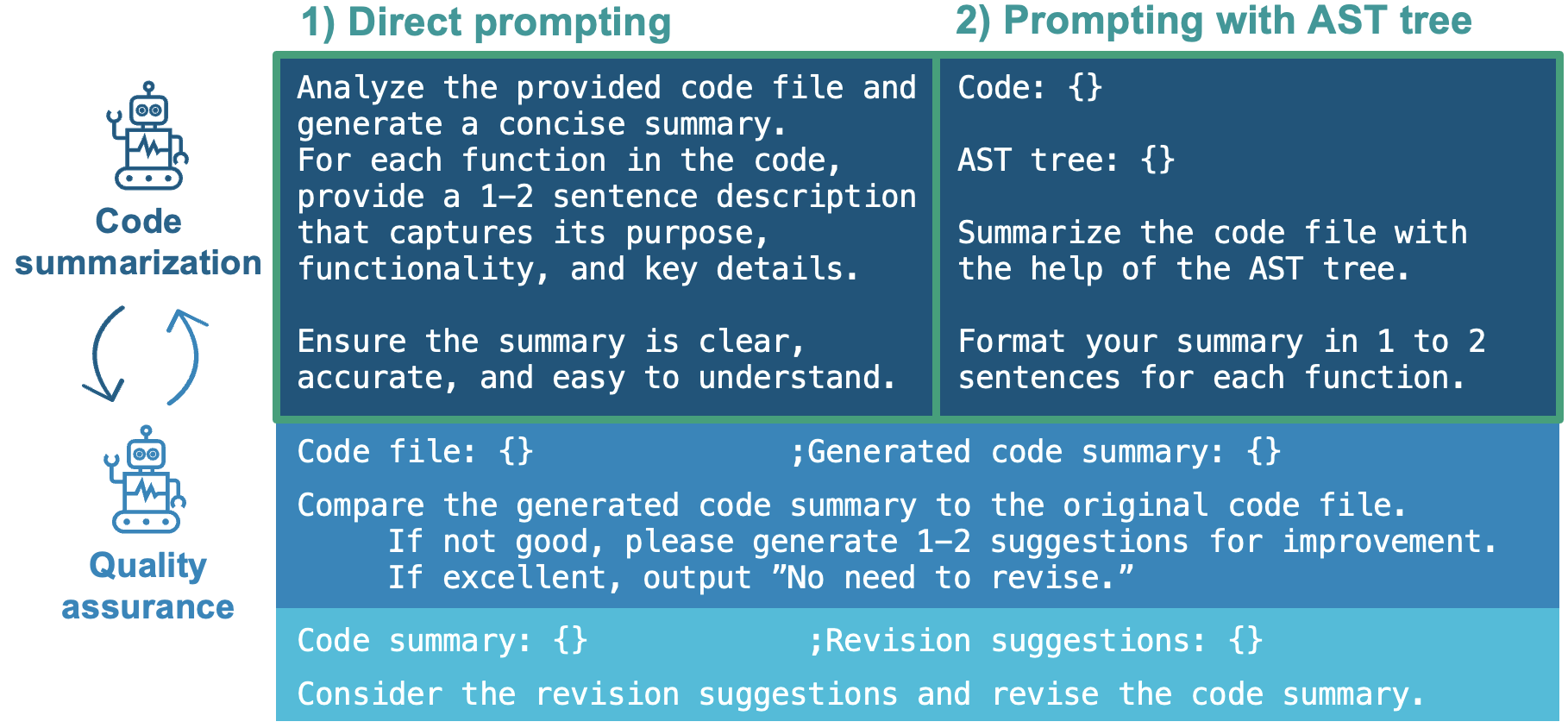}
  \caption{The prompts used to summarize the function-level code files. We experiment with two methods: direct prompting and prompting with AST tree.}
  \label{fig:prompt_code}
  \Description{The prompts used to summarize the function-level code files. We experiment with two methods: direct prompting and prompting with AST tree.}
\end{figure}

Figure \ref{fig:prompt_code} illustrates our two experimental strategies for function-level code summarization: 1) a direct prompt designed by us and 2) a prompt enhanced with the abstract syntax tree (AST) \cite{sun2023abstract} of code. For reference, AST is a hierarchical tree representation of source code that captures its syntactic structure, including elements such as function declarations, variable assignments, and control flow statements. The latter strategy with AST was used in baseline HR-CS \cite{dhulshette2025hierarchical}, and we aim to validate its effectiveness in our experimental context.

In this scenario, Agent4cs comprises two collaborative agents: a code summarization agent and a quality assurance agent. Given a code file, the summarization agent generates a concise summary, which is then analyzed by the quality assurance agent that provides feedback for refinement. Based on this feedback, the summarization agent improves the summary, establishing an effective feedback loop between the agents.

\subsection{Agentic Workflow for Hierarchical Summarization}
For higher-level folder summarization, we follow a hierarchical procedure, progressively aggregating summaries from individual code files to construct a complete repository summary. This method leverages the multi-layered tree structure of folders. 

\begin{figure}[h]
  \centering
  \includegraphics[width=0.8\linewidth]{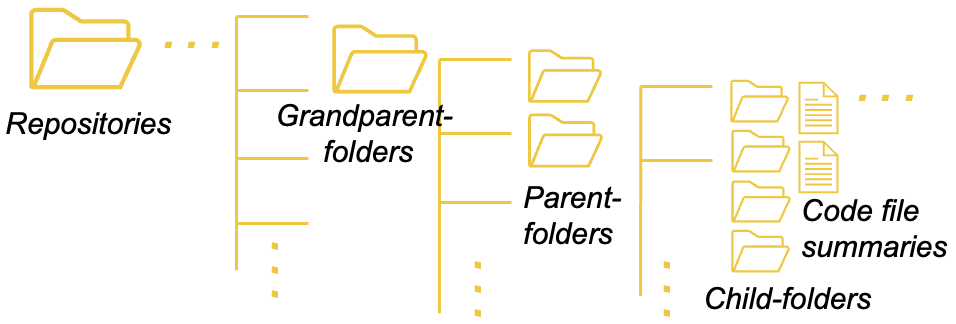}
  \caption{An illustrative example showing the hierarchical relationship between child, parent, and grandparent folders.}
  \label{fig:folder_relation}
  \Description{An illustrative example showing the hierarchical relationship between child, parent, and grandparent folders.}
\end{figure}

\begin{figure}[h]
  \centering
  \includegraphics[width=1\linewidth]{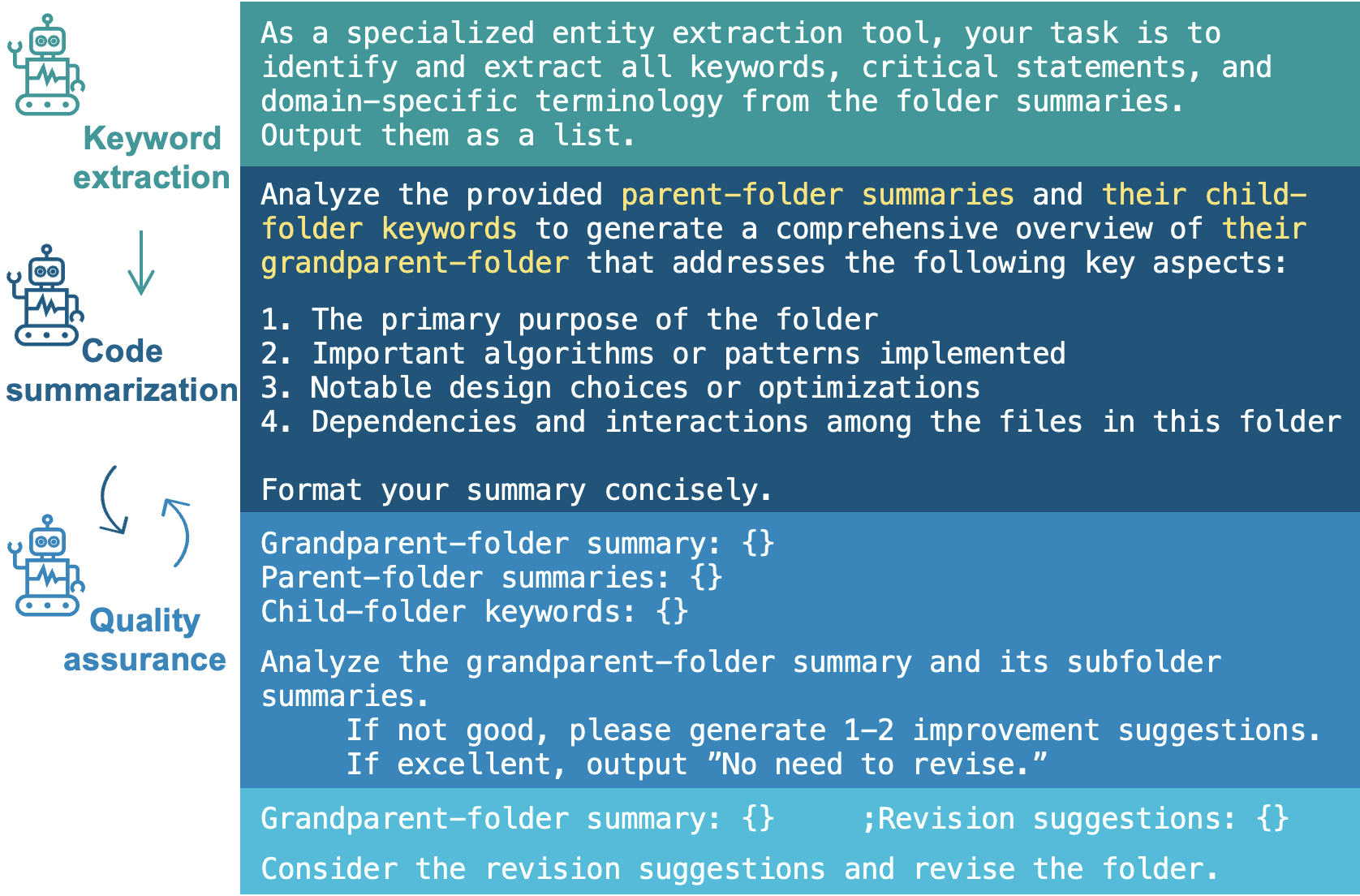}
  \caption{The prompts used to summarize a parent folder based on its folder and subfolder summaries.}
  \label{fig:prompt_folder}
  \Description{The prompts used to summarize a parent folder based on its folder and subfolder summaries.}
\end{figure}

Given the generated function-level summaries for code files at the deepest repository layer, Agent4cs continues to summarize folders at the next layer. Starting from the third layer, i.e.,~the grandparent layer in Figure \ref{fig:folder_relation}, we adopt the prompting strategy in Figure \ref{fig:prompt_folder}. An additional keyword extraction agent is introduced to identify keywords and critical information from child-folder summaries, which are then combined with parent-folder summaries of next layer and passed as input to the summarization agent together. This initial round produces a draft of grandparent-folder summary.
Subsequently, a quality assurance agent reviews the draft and provides suggestions for improvements. Based on this feedback, the summarization agent refines the draft iteratively and generates the finalized version.
Following this hierarchical grandparent-parent-child framework across three neighboring folder layers, the multi-agent system constructs the summaries at each successive layer until reaching the top repository level.

Keyword extraction serves as a critical step to enhance the hierarchical connection between multiple folder layers. Compared to simply using parent-folder summaries, aggregating child-folder information through keywords provides a more condensed representation of the underlying content while maintaining manageable input sizes. Notably, due to context window limitations, including all child-folder summaries directly in the prompt is impossible. Therefore, we implement this keyword-based alternative, enabling the system to leverage lower-level insights within computational constraints.

\section{Experiments}
\subsection{Datasets}

Prior research on folder- and repository-level code summarization faces significant limitations due to the lack of appropriate data and evaluation resources. All public datasets exclusively focus on function-level assessment, and previous studies on repository-level summarization have not open-sourced their datasets. 
To address these constraints, we conduct a comprehensive search of public datasets and identify two that provide detailed function paths within repositories. Using the provided paths, we restore the entire repositories to enable future hierarchical summarization research.

The CodeSearchNet (CSN) corpus \cite{husain2019codesearchnet} 
contains approximately 6 million functions across 6 programming languages (Go, Java, JavaScript, PHP, Python, and Ruby) 
and has been widely used in the study of code summarization \cite{sun2023automatic, fang2024esale, sounthiraraj2025code, sun2025source, shin2025prompt, makharev2025code, sun2024extractive}. 
The CodeXGLUE benchmark \cite{lu1codexglue} is established to advance research in program understanding and generation. It features 10 diverse tasks across 14 code datasets, including four primary categories of code-related challenges: code-code, text-code, code-text, and text-text tasks.

Since our research focuses on hierarchical summarization of large codebases, we selectively include repositories from these two benchmarks by applying specific criteria: The repositories must 1) contain more than 1000 functions with clean comments and 2) have a folder depth exceeding 7 layers. In addition to these two datasets, we add the open-source pybind repository to evaluation. This results in 6 high-quality repositories containing human-written, function-level descriptions (used as groundtruth) for research, as shown in Table \ref{tab:datasets}. Notably, the functions in the collected repositories may share the same folder path, indicating that they are located within the same code file. This explains why the number of files is significantly fewer than the total number of functions. All repositories can be freely accessed online \cite{repositories}.

\begin{table}[!h]
    \caption{The information of collected repositories.}
	\label{tab:datasets}
    \centering
    \resizebox{0.46\textwidth}{!}{
	\begin{tabular}{llllll}\toprule
		\textit{Dataset} & \textit{Repository} & \textit{Domain} & \textit{Depth} & \textit{Code files} & \textit{Functions}  \\ \midrule
		CodeXGlue & base & Robotics & 7 & 216 & 1116 \\ 
		CodeXGlue & coretools & IoT tools & 8 & 347 & 1683 \\
		CSN & twilio-python & Communication & 9 & 293 & 1199 \\
		CSN & turicreate & AI tools & 10 & 237 & 1725 \\
		CSN & pants & Monorepo tools & 10 & 300 & 1186 \\
        by us & pybind & C++ Integration & 13 & 598 & 1319 \\
        \bottomrule
	\end{tabular}
    } 
    
\end{table}

Beyond standard clean code evaluation, we obfuscate the source code -- renaming variables, randomizing identifiers, and stripping documentation, while preserving functional equivalence. 
These experiments help provide insights into whether LLMs truly comprehend code structure and logic flow or merely rely on surface-level patterns. 

\subsection{Language Models}
We strategically select seven LLMs as experimental representatives that span different performance tiers. To fully exploit the capabilities of our agentic framework, we employ four frontier models -- GPT-5 \cite{2025gpt5}, gpt-4.1 \cite{2024gpt41}, GPT-4o\cite{hurst2024gpt4o}, and gemini-2.5-flash \cite{comanici2025gemini} -- accessed via APIs due to computational constraints. Complementing these top-tier models, we incorporate three open-source alternatives that ensure reproducibility and broader applicability: Meta's LLaMA-3.1-8B \cite{dubey2024llama}, Alibaba's Qwen3-8B \cite{yang2025qwen3technicalreport}, and Google's Gemma-3-4B \cite{team2025gemma}. 


\subsection{Metrics}
For function-level summary evaluation with groundtruth, we employ a broad spectrum of metrics: 

\begin{itemize}
    \item [1)] Regarding summary-to-summary text similarity, we apply  \textbf{BLEU-1} \cite{papineni2002bleu} and \mbox{\textbf{ROUGE-L}} \cite{lin2004rouge} for evaluation. 
    BLEU-1 measures the proportion of individual words (unigrams) in the candidate summary that match the reference summary, while ROUGE-L evaluates the longest common subsequence between the candidate and reference texts. 



    
    
    \item [(2)] For summary-to-summary semantic similarity, we apply \textbf{BERTScore} \cite{zhang2019bertscore} and \textbf{SentenceBert} \cite{reimers2019sentence}. 
    BERTScore computes token-level semantic similarity by comparing BERT-based word embeddings between the candidate and reference summaries, while SentenceBert measures sentence-level similarity by encoding summaries into dense vector representations and calculating their cosine similarity.
    
    \item [(3)] Regarding summary-to-code semantic similarity metrics, we apply \textbf{SIDE} \cite{mastropaolo2024evaluating} and \textbf{LLM-as-a-judge} \cite{wu2025can, crupi2025effectiveness}.
    SIDE leverages a fine-tuned encoder model to assess how well the candidate summary captures code semantics, while LLM-as-a-judge employs prompt-based assessment to grade the generated summaries. 
    Figure \ref{fig:llm_as_a_judge} presents the prompt used for LLM-as-a-judge, with grades assigned on a scale from 1 to 4.

\begin{figure}[h]
    \raggedleft  
  \includegraphics[width=0.9\linewidth]{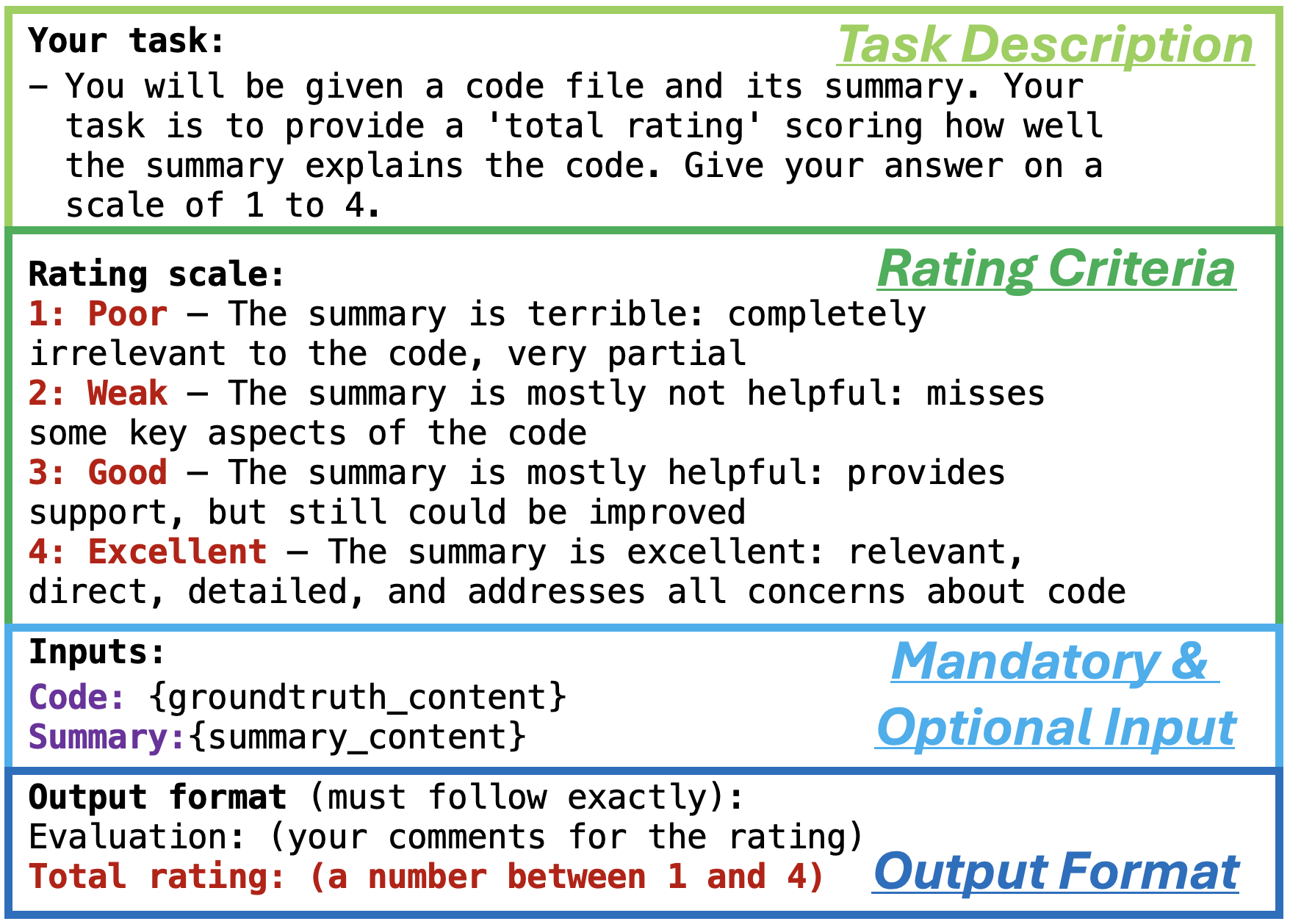}
  \caption{The prompt for LLM-as-a-judge.}
  \label{fig:llm_as_a_judge}
  \Description{The prompt for LLM-as-a-judge.}
\end{figure}

    \item [(4)] We further include human evaluation for comparison. To align with LLM-as-a-judge, three software developers assess summaries independently on a scale from 1 to 4, with final scores averaged for fairness. We selected human evaluators based on their software development experience, code-review skills, and relevant domain expertise. Finally, we chose two senior software engineers, each with three years of development experience, and one junior software developer in industry. 
    In addition to their own experience and judgment criteria, we also provided the prompt and grading schema used in LLM-as-a-judge to ensure consistent evaluation. Their evaluation results served as valuable reference in our experimental context. 

\end{itemize}


For hierarchical summary evaluation, no open-source datasets provide groundtruth annotations for folder-level summaries, making traditional reference-based metrics inapplicable. Consequently, we define the following reference-free metrics for evaluation:

\begin{itemize}

\item [1)] \textbf{Semantic similarity} -- measures the average pairwise cosine similarity between a parent-folder summary and each of its child-folder summaries, computed using SentenceBERT embeddings \cite{reimers2019sentence}. This metric measures the semantic connectivity between parent folders and their children.
        
    \item [2)] \textbf{Keyword coverage rate} -- quantifies the proportion of keywords from child-folder summaries that appear in their parent-folder summary. We extract keywords from child-folder summaries using the Scikit-learn TF-IDF library \cite{scikit-learn}, then compute and average the coverage rates across parent folders at each hierarchical depth. 

    \item [3)] \textbf{Normalized keyword coverage rate} -- represents the keyword coverage rate normalized by summary length.  This metric assesses whether high keyword coverage is achieved efficiently, without introducing unnecessary redundancy through overly verbose summaries. 
    Equation \ref{equ:ratio}, \ref{equ:max}, and \ref{equ:normalization} demonstrate the formula for normalization, where $R_i$ stands for the ratio between keyword coverage rate $r_i$ and summary length $l_i$ for the $i$-th selected model. 
    
    \begin{eqnarray}
    \label{equ:ratio} 
    R_i & = & \frac{r_i}{l_i} \\  R_{max} & =  &max (\frac{r_i}{l_i})
    \label{equ:max}
    \end{eqnarray}
    
    Given the maximum ratio, here is the formula to compute normalized keyword coverage rate: 
    \begin{eqnarray}
    \label{equ:normalization}
    R_{Ni}  & = & \frac{R_i}{R_{max} }
    \end{eqnarray}

    \item [4)] \textbf{Readability} -- 
    measures the Flesch reading-ease score \cite{dubay2004principles} of summaries, which considers the number of words per sentence and syllables per word. According to the Flesch–Kincaid framework \cite{dubay2004principles, readability_wiki}, reading-ease scores range from 0 to 100, with higher scores indicating easier readability. For example, scores of 30 to 50 correspond to college entry-level reading difficulty, while scores of 10 to 30 indicate graduate-level complexity. We use the textstat library to calculate the scores for each summary, then average the results for each evaluated LLM.

\end{itemize}

We select these metrics considering multiple facts. Relying solely on semantic similarity is insufficient, as it may reward summaries that simply repeat content from child folders, achieving high coherence while lacking meaningful abstraction and failing to highlight key information. For this reason, we introduce keyword coverage as the second metric to ensure that critical concepts, identities, and function names are preserved in the summary. 
While these two metrics measure whether parent-folder summaries adequately capture child-folder content, normalized keyword coverage rate and readability focus on efficiency and usability, ensuring that the generated summaries achieve high coverage without unnecessary redundancy and remain readable to developers in practice. 
Notably, we do not employ LLM-as-a-judge for hierarchical summary evaluation, as this approach may introduce evaluation bias when assessing summaries generated by the same or similar LLMs. Instead, we prioritize objective and reproducible metrics that maintain consistency across all hierarchical depths.

\subsection{Function-level Code Summarization}


\begin{table*}[ht]  
\caption{Performance comparison of function-level code summarization on selected repositories using text-based, semantic-based, LLM-based metrics, and human evaluation.}  
\label{tab:functional_evaluation}
\centering  
\resizebox{1\linewidth}{!}{
\setlength{\tabcolsep}{4pt}

\begin{tabular}{c c c c c c c c c c c c c}  

\toprule
 &  &  & \multicolumn{4}{c}{Summary to Summary  }  
    & \multicolumn{6}{c}{Code to Summary} \\   

Code quality & Methods & Models & \multicolumn{2}{c}{Text Based} & \multicolumn{2}{c}{Semantic Based}  
&  & \multicolumn{4}{c}{LLM-as-a-judge} & Human\\  

\cmidrule(lr){4-5}  \cmidrule(lr){6-7} \cmidrule(lr){9-12} \cmidrule(lr){13-13}

  & & & \multicolumn{1}{c}{Rouge-L} & \multicolumn{1}{c}{BLEU-1} & BERTScore (F1) & Sentence-BERT & \multicolumn{1}{c}{SIDE} & GPT-4o & GPT-4.1 & GPT-5 & Gemini-2.5-flash & \\ 

\midrule  

&  &GPT-4o & \textbf{0.189} & 0.184 & \textbf{0.826} & 0.624 & 0.864 & 3.82 & 3.97 & 3.34 & 3.95 & 3.52\\  

&  & GPT-4.1 & 0.177 & 0.189 & 0.822 & \textbf{0.672} & \textbf{0.880} & 3.75 & 3.99 & 3.37 & 3.96 & 3.61\\  

&  & GPT-5 & 0.171 & \textbf{0.197} & 0.824 & 0.650 & 0.778 & \textbf{3.92} & \textbf{4.00} & \textbf{3.92} & \textbf{4.00} & \textbf{3.72}\\                                   
& Agent4cs & Gemini-2.5-flash & 0.172 & 0.164 & 0.822 & 0.647 & 0.880 & 3.62 & 3.94 & 3.43 & 3.96 & 3.58\\  
&  & LLaMA-3.1-8B  & 0.151 & 0.151 & 0.815 & 0.627 & 0.772 & 3.32 & 3.71 & 2.93 & 3.66 & 3.18\\  
&  &Qwen3-8B  & 0.114 & 0.118 & 0.811 & 0.588 & 0.792 & 3.74 & 3.98 & 3.84 & 3.95 & 1.92\\  
&  &Gemma3-4B  & 0.145 & 0.143 & 0.812 & 0.606 & 0.782 & 3.47 & 3.82 & 2.74 & 3.81 & 1.93\\

\cline{3-13} 

&  &GPT-4o & \textbf{0.177} & 0.176 & 0.821 & 0.605 & \textbf{0.879} & 3.76 & 3.89 & 3.28 & 3.94 & 3.52\\   
                     
&  & GPT-4.1  & 0.172 & 0.179 & \textbf{0.822} & \textbf{0.656} & 0.812 & 3.64 & 3.97 & 3.29 & 3.94 & 3.60\\
                    
Clean & Baseline & GPT-5  & 0.166 & \textbf{0.193} & 0.810 & 0.624 & 0.768 & \textbf{3.89} & \textbf{3.99} & \textbf{3.88} & \textbf{4.00} & \textbf{3.64}\\

code & HR-CS  & Gemini-2.5-flash  & 0.170 & 0.166 & 0.812 & 0.622 & 0.832 & 3.53 & 3.92 & 3.41 & 3.96 & 3.51\\
                                 
&  & LLaMA-3.1-8B   & 0.144 & 0.149 & 0.818 & 0.623 & 0.792 & 3.24 & 3.54 & 2.81 & 3.62 & 3.03\\

&  &Qwen3-8B  & 0.113 & 0.112 & 0.803 & 0.586 & 0.783 & 3.69 & 3.94 & 3.78 & 3.92 & 2.68\\                  
&  &Gemma3-4B  & 0.140 & 0.147 & 0.803 & 0.601 & 0.772 & 3.38 & 3.79 & 2.99 & 3.75 & 1.90\\

\cline{3-13} 

&  &GPT-4o & \textbf{0.181} & 0.179 & 0.822 & 0.615 & 0.859 & 3.79 & 3.94 & 3.28 & 3.95 & 3.53\\   
                     
&  & GPT-4.1  & 0.166 & 0.176 & 0.814 & 0.651 & \textbf{0.864} & 3.65 & 3.98 & 3.28 & 3.93 & \textbf{3.60}\\                      
& Baseline & GPT-5  & 0.161 & \textbf{0.185} & \textbf{0.826} & \textbf{0.666} & 0.774 & \textbf{3.89} & \textbf{4.00} & \textbf{3.90} & \textbf{4.00} & 3.60\\  
& CS-BF  & Gemini-2.5-flash  & 0.172 & 0.162 & 0.807 & 0.617 & 0.836 & 3.54 & 3.93 & 3.44 & 3.97 & 3.58\\                                   
&  & LLaMA-3.1-8B   & 0.149 & 0.152 & 0.826 & 0.618 & 0.783 & 3.27 & 3.56 & 2.83 & 3.66 & 3.23\\  
&  & Qwen3-8B  & 0.118 & 0.109 & 0.798 & 0.590 & 0.787 & 3.72 & 3.95 & 3.81 & 3.90 & 2.59\\                    
&  & Gemma3-4B  & 0.135 & 0.145 & 0.810 & 0.595 & 0.779 & 3.36 & 3.81 & 3.01 & 3.77 & 2.03\\

\cline{2-13}

&  & GPT-4o & 0.161 & 0.168 & \textbf{0.819} & 0.530 & 0.762 & 3.35 & 3.90 & 3.08 &  3.88 & -\\  
&  & GPT-4.1& 0.157 & 0.173 & 0.817 & \textbf{0.671} & \textbf{0.789} & 3.53 & 3.96 & 3.20 & 3.93 & -\\
&  & GPT-5  & 0.154 & \textbf{0.182} & 0.819 & 0.612 & 0.704 & \textbf{3.88} & \textbf{3.99} &  \textbf{3.84} &  \textbf{4.00} & -\\
& Agent4cs  & Gemini-2.5-flash  & \textbf{0.162} & 0.156 & 0.792 & 0.622 & 0.742 & 3.35 & 3.90 & 3.23 & 3.91 & -\\
&  & LLaMA-3.1-8B   & 0.144 & 0.139 & 0.808 & 0.613 & 0.708 & 2.98 & 3.48 & 2.78 & 3.46 & -\\
&  &Qwen3-8B  & 0.103 & 0.102 & 0.794 & 0.566 & 0. 732& 3.39 & 3.89  & 3.12 & 3.87 & -\\                  
&  &  Gemma3-4B  & 0.130 & 0.132 & 0.790 & 0.578 & 0.707 & 3.24 & 3.63 & 2.84 & 3.58 & -\\

\cline{3-13} 
                    
&  & GPT-4o  & 0.150 & 0.159 & 0.814 & 0.509 & \textbf{0.763} & 3.34 & 3.88 & 3.02 & 3.86 & -\\  

&  & GPT-4.1& 0.149 & 0.163 & \textbf{0.815} & \textbf{0.623} & 0.723 & 3.31 & 3.96 & 3.10 & 3.92 & -\\    
                       
Obfuscated & Baseline & GPT-5  & \textbf{0.156} & \textbf{0.181} & 0.809 & 0.599 & 0.698 & \textbf{3.86} & \textbf{3.99} & \textbf{3.84} &  \textbf{3.99} & -\\    
                       
code &  HR-CS & Gemini-2.5-flash & 0.149 & 0.159 & 0.798 & 0.614 & 0.728 & 3.25 & 3.87  & 3.01 &    3.85  & - \\

&  & LLaMA-3.1-8B   & 0.140 & 0.133 & 0.799 & 0.603 & 0.679 & 3.12 & 3.58 & 2.76 & 3.49 & -\\                            

&  &Qwen3-8B  & 0.105 & 0.108 & 0.786 & 0.571 & 0.723 & 3.24 & 3.82  & 3.07 & 3.86 & -\\ 

&  & Gemma3-4B  & 0.129 & 0.129 & 0.787 & 0.570 & 0.688 & 3.18 & 3.54 & 2.68 & 3.49  & -\\

\cline{3-13}

&  & GPT-4o & 0.156 & 0.173 & \textbf{0.816} & 0.535 & 0.770 & 3.37 & 3.88 & 3.10 & 3.86 & -\\    
&  & GPT-4.1& 0.152 & 0.179 & 0.812 & \textbf{0.645} & \textbf{0.775} & 3.55 & 3.94 & 3.22 & 3.91 & -\\      
& Baseline & GPT-5  & \textbf{0.160} & \textbf{0.179} & 0.813 & 0.607 & 0.699 & \textbf{3.86} & \textbf{3.99} & \textbf{3.81} & \textbf{3.99} & -\\  
& CS-BF  & Gemini-2.5-flash  & 0.160 & 0.153 & 0.788 & 0.618 & 0.736 & 3.38 & 3.92 & 3.21 & 3.92 & -\\   
&  & LLaMA-3.1-8B   & 0.149 & 0.136 & 0.813 & 0.609 & 0.703 & 3.01 & 3.46 & 2.76 & 3.49 & -\\   
&  & Qwen3-8B  & 0.108 & 0.107 & 0.789 & 0.562 & 0.727 & 3.41 & 3.91 & 3.14 & 3.89 & -\\                    
&  & Gemma3-4B  & 0.127 & 0.137 & 0.795 & 0.580 & 0.712 & 3.26 & 3.66 & 2.86 & 3.56 & -\\

\bottomrule  

\end{tabular}  
} 

\end{table*}  

Table \ref{tab:functional_evaluation} demonstrates the evaluation outcomes for summarizing functional-level code files. 
Agent4cs surpasses the two baselines in all considered metrics.
Regarding traditional evaluation metrics, no individual model emerges as a clear winner, while the selected GPT models consistently demonstrate superior performance across all scenarios, achieving 0.189 in Rouge-L, 0.197 in BLEU-1, 0.826 in BERT F1 Score, and 0.672 in Sentence-BERT similarity to groundtruth. Furthermore, our experiments with obfuscated code reveal that LLMs can comprehend obfuscated code effectively. Despite a minimal performance drop compared to clean code scenarios, the summaries maintain readable and informative in practice to provide developers with meaningful project insights.

In case of using LLM-as-a-judge, GPT-5 performs best, consistently achieving scores above 3.81 across all judge models, and occasionally securing perfect 4.0 ratings from several evaluators. Another interesting finding emerges with Qwen: while showing poor results on traditional metrics, it achieves significantly better performance under LLM-as-a-judge evaluation. This improvement can likely be attributed to its post-training strategy, which incorporates distilled output logits from teacher models \cite{yang2025qwen3technicalreport}. The enhanced performance potentially stems from this distilled knowledge, which aligns Qwen more closely with flagship models like GPT-5 and Gemini-2.5-flash.

Furthermore, behavior analysis on judge models reveals distinct evaluation patterns: GPT-4.1 tends to score other models highly, potentially due to its prioritization of code intelligence over language understanding and quality assessment, leading to favorable evaluation of LLM-generated content. In contrast, GPT-5 applies more rigorous evaluation criteria, often assigning average scores below 3.0 to compact models. This observation indicates that models with superior language comprehension capabilities tend to evaluate more strictly in summarization tasks. 

\begin{figure}[htbp]
  \centering
  \includegraphics[width=0.58\linewidth]{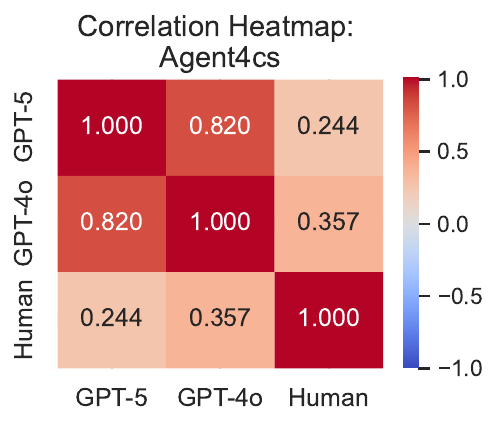}
  \caption{The correlation heatmap between human and LLM-as-a-judge when using Agent4cs.}
  \label{fig:correlation_human_llms}
  \Description{The correlation heatmap between human and LLM-as-a-judge when using Agent4cs.}
\end{figure}

To investigate the alignment between LLMs and human developers, we compare their scores by computing Pearson correlation coefficients, which ranges from \textminus{1} to 1, with higher positive values indicating stronger agreement. We exclude GPT-4o and gemini-2.5-flash from this analysis due to their consistently high scoring patterns observed in Figure \ref{tab:functional_evaluation}. As shown in Figure \ref{fig:correlation_human_llms}, LLMs and human judges exhibit positive correlation for the Agent4CS approach, indicating that LLM-as-a-judge can reasonably approximate human judgment in our experimental context. 

\subsection{Hierarchical Code Summarization}
\begin{table*}[ht]  

\caption{The performance of Agent4cs and the two baselines at each depth level, measured by semantic similarity score and keyword coverage rate. } 
\label{tab:hierarchical}

\centering  
\resizebox{1\textwidth}{!}{
\setlength{\tabcolsep}{2.5pt}
\begin{tabular}{c c | c c c | c c c | c c c | c c c | c c c | c c c | c c c }  
\toprule

 &  & \multicolumn{3}{c}{\textbf{GPT-4o}} & \multicolumn{3}{c}{\textbf{GPT-4.1}} & \multicolumn{3}{c}{\textbf{GPT-5}} & \multicolumn{3}{c}{\textbf{Gemini-2.5-flash}} & \multicolumn{3}{c}{\textbf{LLaMA-3.1-8B}} & \multicolumn{3}{c}{\textbf{Qwen3-8B}} & \multicolumn{3}{c}{\textbf{Gemma3-4B}} \\  

\cmidrule(lr){3-5}  \cmidrule(lr){6-8} \cmidrule(lr){9-11}
\cmidrule(lr){12-14}  \cmidrule(lr){15-17} \cmidrule(lr){18-20} \cmidrule(lr){21-23}

\textbf{Metrics} & \textbf{Depth} & 
\textbf{CS-BF} & \textbf{HR-CS} & \textbf{Agen.} & \textbf{CS-BF} & \textbf{HR-CS} & \textbf{Agen.}& \textbf{CS-BF} & \textbf{HR-CS} & \textbf{Agen.}& \textbf{CS-BF} & \textbf{HR-CS} & \textbf{Agen.}& \textbf{CS-BF} & \textbf{HR-CS} & \textbf{Agen.}& \textbf{CS-BF} & \textbf{HR-CS} & \textbf{Agen.}& \textbf{CS-BF} & \textbf{HR-CS} & \textbf{Agen.} \\  

\midrule
&Layer 1&-&0.583&0.576&-&0.611&0.594&-&0.633&0.701&-&0.533&0.535&-&0.622&0.575&-&0.480&0.537&-&0.251&0.321\\
&Layer 2&-&0.584&0.572&-&0.623&0.607&-&0.647&0.720&-&0.542&0.592&-&0.638&0.591&-&0.488&0.603&-&0.316&0.332\\
&Layer 3&-&0.594&0.582&-&0.629&0.609&-&0.665&0.715&-&0.524&0.540&-&0.690&0.575&-&0.528&0.638&-&0.482&0.731\\
&Layer 4&-&0.569&0.599&-&0.635&0.620&-&0.689&0.744&-&0.519&0.581&-&0.697&0.611&-&0.516&0.581&-&0.452&0.555\\
&Layer 5&-&0.617&0.653&-&0.648&0.650&-&0.714&0.763&-&0.550&0.702&-&0.712&0.664&-&0.516&0.608&-&0.446&0.423\\
Semantic &Layer 6&-&0.621&0.660&-&0.663&0.676&-&0.731&0.767&-&0.577&0.670&-&0.720&0.672&-&0.529&0.601&-&0.473&0.437\\
&Layer 7&-&0.660&0.682&-&0.683&0.692&-&0.742&0.811&-&0.663&0.683&-&0.694&0.698&-&0.595&0.708&-&0.345&0.328\\
rate&Layer 8&-&0.659&0.737&-&0.730&0.693&-&0.768&0.836&-&0.626&0.752&-&0.769&0.740&-&0.608&0.685&-&0.372&0.510\\
&Layer 9&-&0.668&0.757&-&0.731&0.700&-&0.786&0.843&-&0.636&0.790&-&0.770&0.714&-&0.621&0.682&-&0.321&0.509\\
&Layer 10&-&0.710&0.776&-&0.741&0.693&-&0.805&0.853&-&0.657&0.801&-&0.790&0.726&-&0.633&0.693&-&0.305&0.382\\
&Layer 11&-&0.754&0.772&-&0.751&0.707&-&0.825&0.862&-&0.709&0.737&-&0.792&0.741&-&0.675&0.760&-&0.282&0.251\\
&Layer 12&-&0.738&0.746&-&0.741&0.677&-&0.823&0.848&-&0.780&0.690&-&0.771&0.707&-&0.671&0.732&-&0.287&0.351\\
&Layer 13&-&0.708&0.801&-&0.729&0.742&-&0.811&0.857&-&0.712&0.776&-&0.738&0.689&-&0.613&0.700&-&0.339&0.588\\
\cline{2-23}

\multicolumn{2}{c}{\textbf{\hspace{0.8cm}Average}}&\multicolumn{1}{|c}{0.651}&0.632&\textbf{0.681}&0.667&\textbf{0.682}&0.672&0.721&0.738&\textbf{0.794}&0.585&0.595&\textbf{0.696}&0.714&\textbf{0.727}&0.683&0.549&0.564&\textbf{0.644}&0.412&0.402&\textbf{0.460}\\

\cline{1-23}

&Layer 1&-&0.594&0.751&-&0.656&0.666&-&0.668&0.718&-&0.592&0.612&-&0.798&0.779&-&0.780&0.785&-&0.447&0.651\\
&Layer 2&-&0.605&0.769&-&0.679&0.677&-&0.668&0.748&-&0.567&0.533&-&0.780&0.745&-&0.784&0.782&-&0.418&0.549\\
&Layer 3&-&0.611&0.782&-&0.659&0.691&-&0.697&0.768&-&0.544&0.561&-&0.800&0.753&-&0.833&0.808&-&0.674&0.866\\
&Layer 4&-&0.614&0.796&-&0.659&0.671&-&0.707&0.768&-&0.557&0.573&-&0.820&0.778&-&0.811&0.812&-&0.646&0.815\\
Keyword&Layer 5&-&0.625&0.804&-&0.656&0.705&-&0.718&0.790&-&0.549&0.559&-&0.833&0.817&-&0.816&0.833&-&0.628&0.739\\
&Layer 6&-&0.631&0.813&-&0.664&0.720&-&0.738&0.802&-&0.588&0.573&-&0.837&0.835&-&0.835&0.795&-&0.640&0.882\\
coverage&Layer 7&-&0.682&0.832&-&0.701&0.720&-&0.764&0.824&-&0.615&0.590&-&0.846&0.847&-&0.895&0.911&-&0.422&0.486\\
&Layer 8&-&0.701&0.859&-&0.722&0.746&-&0.775&0.861&-&0.685&0.650&-&0.872&0.899&-&0.933&0.942&-&0.443&0.559\\
rate&Layer 9&-&0.702&0.867&-&0.741&0.768&-&0.787&0.870&-&0.625&0.684&-&0.872&0.884&-&0.940&0.924&-&0.379&0.578\\
&Layer 10&-&0.725&0.873&-&0.759&0.755&-&0.785&0.873&-&0.673&0.668&-&0.895&0.886&-&0.949&0.920&-&0.360&0.496\\
&Layer 11&-&0.754&0.868&-&0.783&0.772&-&0.818&0.893&-&0.696&0.716&-&0.889&0.887&-&0.962&0.989&-&0.343&0.552\\
&Layer 12&-&0.759&0.858&-&0.762&0.777&-&0.817&0.903&-&0.753&0.750&-&0.890&0.898&-&0.957&0.941&-&0.311&0.434\\
&Layer 13&-&0.715&0.870&-&0.722&0.767&-&0.842&0.890&-&0.703&0.658&-&0.878&0.831&-&0.957&0.923&-&0.428&0.589\\
\cline{2-23}
\multicolumn{2}{c}{\textbf{\hspace{0.8cm}Average}}&\multicolumn{1}{|c}{0.682}&0.659&\textbf{0.829}&0.686&0.691&\textbf{0.722}&0.732&0.748&\textbf{0.809}&0.582&0.603&\textbf{0.635}& \textbf{0.846}&0.827&0.843&0.841&\textbf{0.872}&0.870&0.543&0.526&\textbf{0.643}\\
  
\bottomrule               

\end{tabular}  
} 

\end{table*}

Table \ref{tab:hierarchical} presents the evaluation results across multiple folder layers, from repository level to bottom file level, showing semantic similarity and keyword coverage rates for all selected LLMs.
The baseline CS-BF follows its own function–class–repository summarization scheme, which is incompatible with our layer-wise folder analysis. Accordingly, we report its overall average score directly in Table \ref{tab:hierarchical}.

With respect to semantic similarity,  
Agent4cs outperforms the two baselines in 5 out of 7 selected models. GPT-5 achieves the best performance with a weighted average score of 0.794 across all folder summaries and layers, while Gemini-2.5-flash shows the largest improvement at over 10\%, followed by Qwen3-8B at 8\%. 
In comparison, LLaMA-3.1-8B and GPT-4.1 are the two underperforming models, with the latter being optimized for code intelligence, which may compromise its language comprehension capabilities.
Furthermore, we notice that the similarity scores gradually decrease from lower to higher hierarchical layers across all models, with degradation ranging from 10\% to 20\%. This observation aligns with our expectations, as higher-level summarization requires capturing more abstract concepts within a constrained word limit, making the task 
more challenging.

Consistent with this trend in semantic similarity, the keyword coverage rate also declines gradually across layers, which reflects the increasing difficulty of capturing all key information as the number of folders grow exponentially. 
Compared to the two baselines, Agent4cs effectively improves the keyword coverage rate in five models. GPT-4o exhibits the most significant enhancement, increasing from 0.682 to 0.829. And for the two underperforming models, 
Agent4cs also achieves on-par performance to baselines with differences of only 0.3\% and 0.2\% respectively. 
An unexpected finding is that compact models lead in performance: Qwen3-8B achieves 0.872, while LLaMA-3.1-8B reaches 0.846. This observation suggested the presence of potential confounding factors. 
Further analysis revealed that token length was the underlying cause.

\begin{figure}[h]
\centering
  \includegraphics[width=0.8\linewidth]{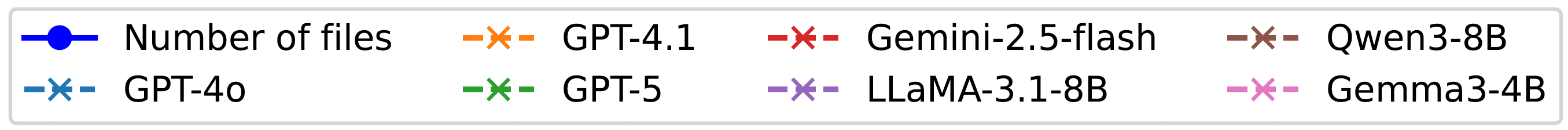}
  \includegraphics[width=1\linewidth]{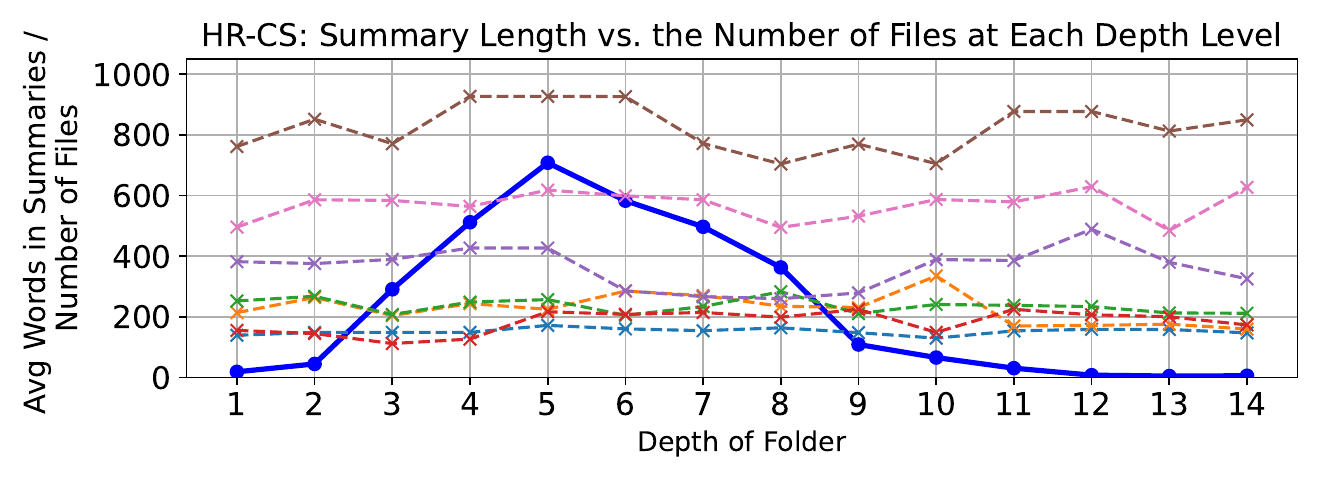}
  \includegraphics[width=1\linewidth]{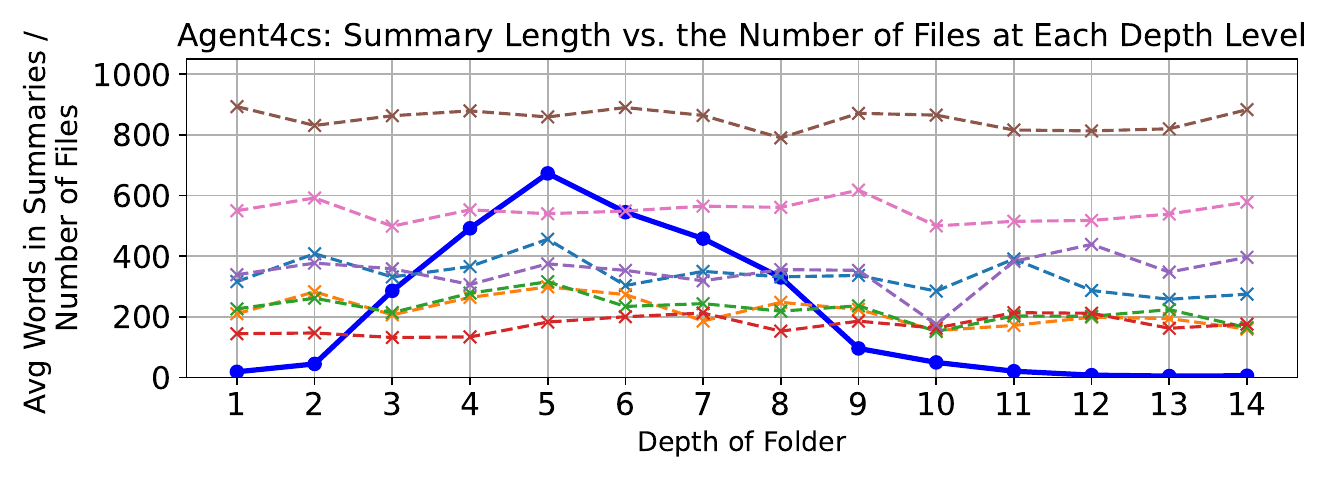}
  \caption{Average summary length of different LLMs vs. the corresponding file count at each depth, while using HR-CS baseline (above) and Agent4cs (below). CS-BF is not included for this layer-based analysis because of its unique summarization scheme. }
  \label{fig:summary length}
  \Description{Average summary length of different LLMs vs. the corresponding file count at each depth, while using HR-CS baseline (above) and Agent4cs (below). CS-BF is not included for this layer-based analysis because of its unique summarization scheme. }
  
\end{figure}

As illustrated in Figure \ref{fig:summary length}, different LLMs produce summaries of varying length. Despite the fluctuating number of files at each depth layer, the summary length generated by each specific LLM is relatively consistent. For large-scale models including GPT-5, GPT-4.1, and Gemini-2.5-flash, regardless of whether Agent4cs or baseline is used, 
the generated summaries contain approximately 200 words. 
In contrast, compact models exhibit greater variation in summary length and tend to produce more redundant outputs.
For instance, Qwen3-8B generates summaries averaging over 800 words while Gemma3-4B produces about 600 words, which explains the reason for high keyword coverage rate in evaluation. Notably, GPT-4o is the only model that performs differently across the two approaches: its summary length increases from 200 to over 350 words when using Agent4cs, leading to more verbose responses.

\begin{figure}[h]
  \centering
  \includegraphics[width=1\linewidth]{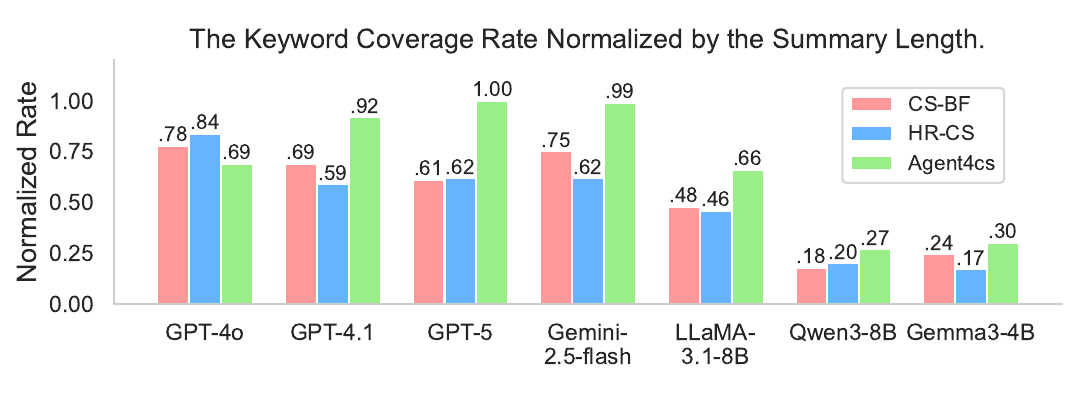}
  \caption{The normalized keyword coverage rate considering the summary length.}
  \label{fig:keywords_length_ratio}
  \Description{The normalized keyword coverage rate considering the summary length.}
\end{figure}

To obtain fair conclusions, we normalize the keyword coverage rate for all selected models based on their summary length and present the results in Figure \ref{fig:keywords_length_ratio}. GPT-5 achieves the highest rate when using Agent4cs, indicating that the model tends to produce more condensed summaries that efficiently incorporate crucial information. The Gemini-2.5-flash secures the second place, closely following the most capable GPT model with a normalized coverage rate of 99\%. This strong performance is largely attributed to having the shortest answers among all models deployed in our agentic system. In general, large-scale models outperform the selected compact models, showing superior ability to produce concise yet informative summaries. In particular, Agent4cs surpasses the baselines in 6 out of 7 evaluated models, with GPT-5 achieving a noteworthy improvement of 38\%. GPT-4o represents the only exception to this trend, which may stem from its earlier release in May 2024 and potentially less mature capability to handle large-scale contexts with feedback loops, leading to increased summary length.

\begin{figure}[h]
  \centering
  \includegraphics[width=1\linewidth]{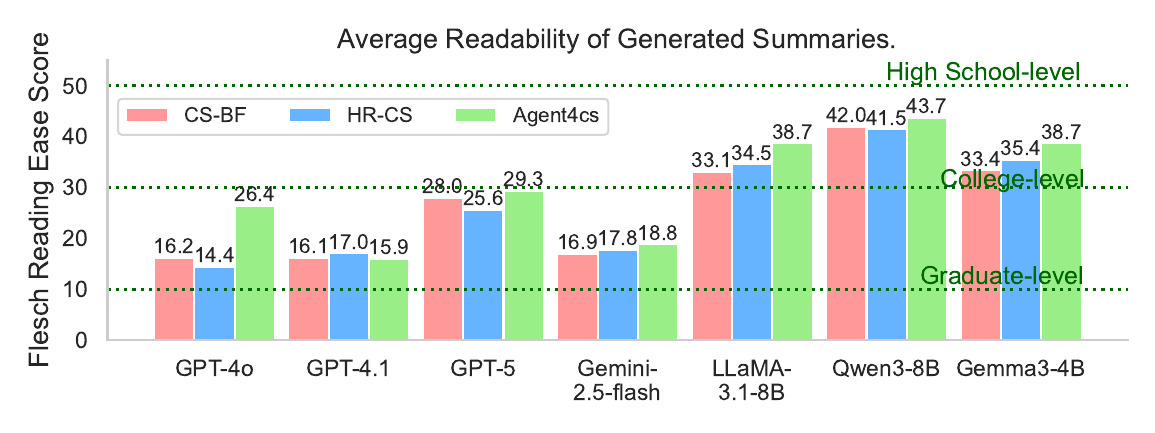}
  \caption{Average Flesch reading-ease scores for summaries generated by different models.}
  \label{fig:readability}
  \Description{Average Flesch reading-ease scores for summaries generated by different models.}
\end{figure}

Furthermore, we assess the summary readability to make sure that they are easy to understand and accessible to developers. Figure \ref{fig:readability} illustrates the average Flesch reading-ease scores for the summaries generated by all included models. 
While large-scale models reach 14 to 30, corresponding to graduate-level complexity, compact models perform better with scores ranging from 33 to 44.
Nevertheless, we should interpret these absolute scores within the context of software engineering. Code summaries inherently contain technical elements such as lengthy identifiers, function names, and explanations of complex algorithms, which naturally lower Flesch reading-ease scores. For developers with domain expertise, this increased complexity should not pose any significant comprehension barrier. 
Agent4cs demonstrates improved readability over the baseline in 6 out of 7 models, with GPT-4.1 as the only exception. This deviation likely reflects the model's emphasis on code intelligence at the expense of natural language generation.

\section{Conclusions and Future Work}
The proposed Agent4cs extends code summarization beyond function level to investigate hierarchical summarization of vast repositories. 
Compared to the baseline, it effectively preserves information when scaling from bottom-level code files to repository-level summaries layer by layer, demonstrating significant improvement in semantic comprehensiveness, normalized keyword average rate, and Flesch-Kincaid readability. 
In principle, modern general-purpose LLMs, such as GPT-5, benefit most from this multi-agent workflow, while compact or specialized models may derive fewer advantages from this loop due to their comparatively limited natural language generation and reasoning capabilities. Furthermore, the empirical study on LLMs' behavior reveals that smaller models tend to produce redundant summaries, while their larger counterparts generate more concise yet informative summaries given the same instructions. Notably, large-scale models with strengthened language capabilities, specifically GPT-5, can also judge LLM-generated content effectively by applying more rigorous rating criteria. These findings highlight the great potential of LLMs and novel agentic solutions in code summarization research.

This work opens several promising avenues for future research. First, while we introduce four metrics for hierarchical code summarization, developing more robust and comprehensive evaluation frameworks remains critical. Furthermore, beyond the dataset released in this work, the community requires larger, more diverse repository-level datasets with high-quality annotated groundtruth summaries. 
Moreover, building on Agent4cs, we plan to explore alternative architectures, incorporating specialized agents to optimize performance while maintaining the current computational efficiency. Finally, as LLMs continue evolving, benchmarking new models within this agentic framework will provide valuable insights into how model capability enhancement translates to improved summarization performance.

\begin{acks}
This work was carried out within the ITEA 4 project GENIUS, as part of the ITEA program, the Eureka Cluster on software innovation. This work was funded by the German Federal Ministry of Research, Technology and Space (BMFTR).
\end{acks}

\bibliographystyle{ACM-Reference-Format}
\bibliography{sample}

\appendix









\end{document}